\documentclass[letterpaper]{article} % DO NOT CHANGE THIS
\usepackage{aaai2026}  % DO NOT CHANGE THIS
\usepackage{times}  % DO NOT CHANGE THIS
\usepackage{helvet}  % DO NOT CHANGE THIS
\usepackage{courier}  % DO NOT CHANGE THIS
\usepackage[hyphens]{url}  % DO NOT CHANGE THIS
\usepackage{graphicx} % DO NOT CHANGE THIS
\usepackage{multirow} 
\usepackage{xcolor}   
\definecolor{darkgreen}{RGB}{20, 100, 0} 
\usepackage{adjustbox}
\usepackage{natbib}  % DO NOT CHANGE THIS AND DO NOT ADD ANY OPTIONS TO IT
\usepackage{caption} % DO NOT CHANGE THIS AND DO NOT ADD ANY OPTIONS TO IT
\usepackage{algorithm}
\usepackage{algorithmic}
\usepackage{amssymb}
\usepackage{array}
\usepackage{amsmath}
\usepackage{newfloat}
\usepackage{listings}

\DeclareCaptionStyle{ruled}{labelfont=normalfont,labelsep=colon,strut=off} % DO NOT CHANGE THIS
\floatstyle{ruled}
\newfloat{listing}{tb}{lst}{}
\floatname{listing}{Listing}
\nocopyright
\title{WEC-DG: Multi-Exposure Wavelet Correction Method Guided by Degradation Description}
\author {  
	Ming Zhao\textsuperscript{\rm 1,2},  
	Pingping Liu\textsuperscript{\rm 2,3}\thanks{Corresponding author.},  
	Tongshun Zhang\textsuperscript{\rm 2,3},  
	Zhe Zhang\textsuperscript{\rm 2,3}  
}  
\affiliations {  
	\textsuperscript{\rm 1}College of Software, Jilin University\\
	\textsuperscript{\rm 2}Key Laboratory of Symbolic Computation and Knowledge Engineering of Ministry of Education, Jilin University\\
	\textsuperscript{\rm 3}College of Computer Science and Technology, Jilin University\\
	\{mingzhao23, tszhang23, zhezhang23\}@mails.jlu.edu.cn, \{liupp\}@jlu.edu.cn  
} 
\usepackage{bibentry}
\begin{document}

\maketitle

\begin{abstract}
Multi-exposure correction technology is essential for restoring images affected by insufficient or excessive lighting, enhancing the visual experience by improving brightness, contrast, and detail richness. However, current multi-exposure correction methods often encounter challenges in addressing intra-class variability caused by diverse lighting conditions, shooting environments, and weather factors, particularly when processing images captured at a single exposure level. To enhance the adaptability of these models under complex imaging conditions, this paper proposes a \textbf{W}avelet-based \textbf{E}xposure \textbf{C}orrection method with \textbf{D}egradation \textbf{G}uidance (\textbf{WEC-DG}). Specifically, we introduce a degradation descriptor within the Exposure Consistency Alignment Module (ECAM) at both ends of the processing pipeline to ensure exposure consistency and achieve final alignment. This mechanism effectively addresses miscorrected exposure anomalies caused by existing methods' failure to recognize 'blurred' exposure degradation. Additionally, we investigate the light-detail decoupling properties of the wavelet transform to design the Exposure Restoration and Detail Reconstruction Module (EDRM), which processes low-frequency information related to exposure enhancement before utilizing high-frequency information as a prior guide for reconstructing spatial domain details. This serial processing strategy guarantees precise light correction and enhances detail recovery. Extensive experiments conducted on multiple public datasets demonstrate that the proposed method outperforms existing algorithms, achieving significant performance improvements and validating its effectiveness and practical applicability.
\end{abstract}

% Uncomment the following to link to your code, datasets, an extended version or similar.
% You must keep this block between (not within) the abstract and the main body of the paper.

\section{Introduction}
\begin{figure}[t]  
	\begin{flushright} % Aligns the figure to the right  
		\centering  
		\includegraphics[width=0.9\columnwidth]{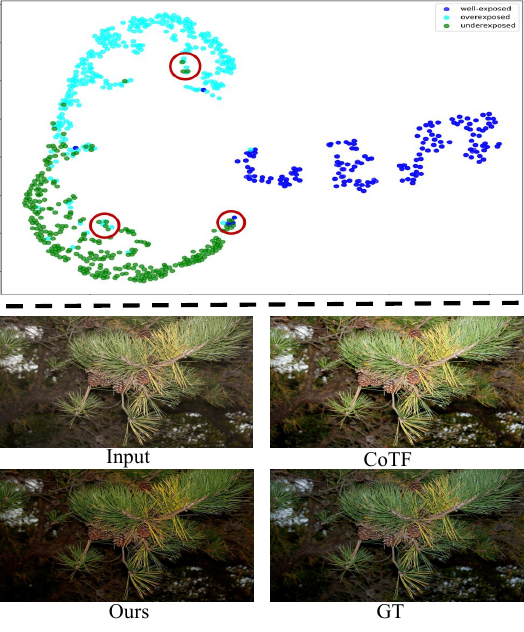} % Adjust width as needed  
		\vspace{-1.2em}   
		\caption{The upper figure presents the t-SNE clustering analysis of images with varying exposure levels, emphasizing outliers that are susceptible to misclassification. The lower figure illustrates an overexposed image that is incorrectly identified as underexposed by the CoTF\citep{li2024real} method, resulting in inappropriate enhancement.}  
		\vspace{-2em}   
		\label{fig1}  
	\end{flushright}   
\end{figure}
In the current digital imaging landscape, acquiring high-quality images is crucial for computer vision and image processing applications\citep{cho2024dual,kim2024beyond, li2024real}. However, real-world image capture often suffers from exposure issues, such as overexposure and underexposure, due to complex lighting, camera settings, and environmental factors. These issues degrade image quality and hinder computational tasks, leading to a significant evolution in exposure research from single-domain low-light enhancement to comprehensive multi-exposure correction.

Low-light enhancement methodologies\citep{chen2018learning, ma2022toward,wang2019underexposed,zhang2024dmfourllie,zhang2025cwnet,zhao2025ref, liu2025multi} have advanced considerably, moving from conventional techniques like histogram equalization and gamma correction to deep learning methods. Pioneering approaches, including Deep-UPE \citep{wang2019underexposed} and Zero-DCE \citep{guo2020zero}, improved local contrast for underexposed images but primarily target underexposure, leaving a gap in addressing multi-exposure degradation. This limitation has spurred the development of broader correction paradigms\citep{afifi2021learning, yang2023implicit, huang2022deep, li2024real, wang2022local}.

Key advancements include the Multi-Scale Exposure Correction (MSEC) method \citep{afifi2021learning}, which simultaneously corrects overexposed and underexposed regions, and the Fourier-based Exposure Correction Network (FECNet) \citep{huang2022deep}, which employs a frequency domain approach to enhance multi-exposure correction through separate amplitude and phase subnetworks.
\begin{figure}[t]
	\centering
	\includegraphics[width=1\columnwidth]{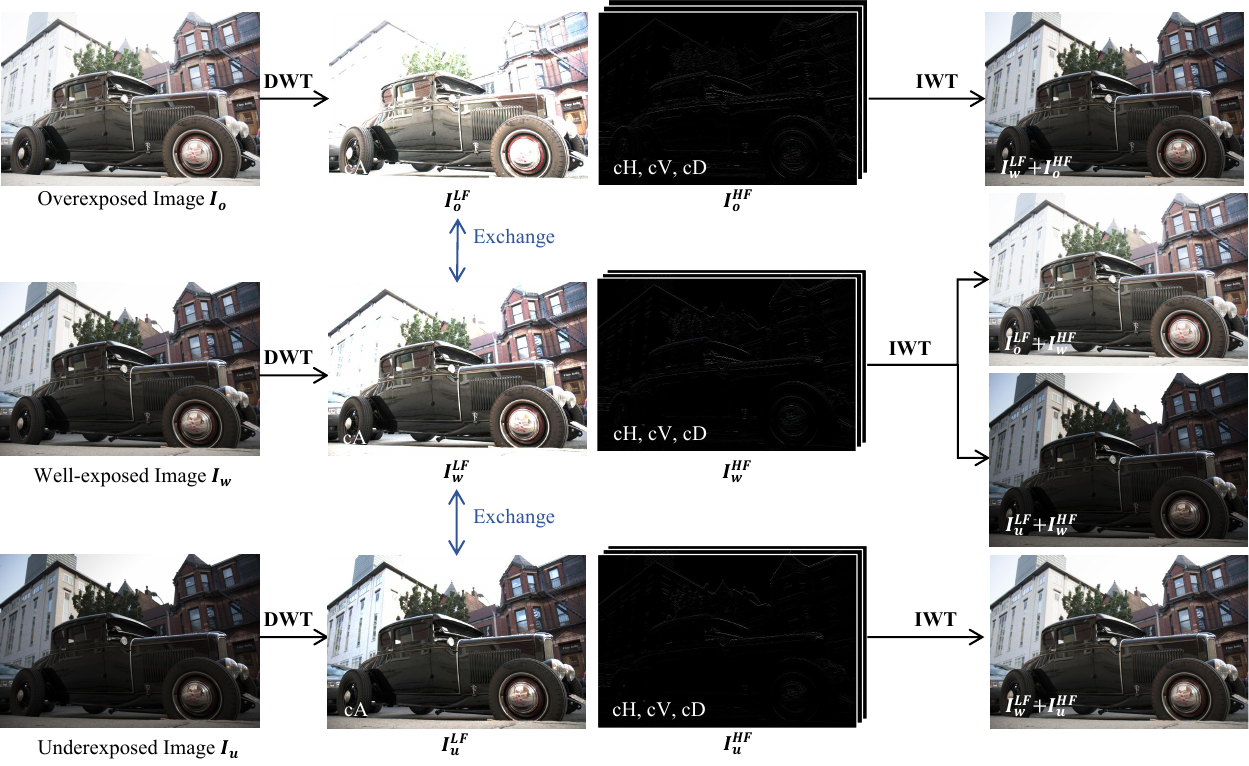} 
	\vspace{-1.5em}
	\caption{Motivational Observation: This figure demonstrates the decoupling of low-frequency (LF) components for illumination and high-frequency (HF) components for detail in multi-exposure images using wavelet transform. By exchanging and reconstructing these components, we can manipulate illumination and detail independently, where LF influences brightness and HF preserves detail with minimal luminance effect.}
	\vspace{-1.5em}
	\label{fig2}
\end{figure}

Despite these advances, current multi-exposure correction approaches face several critical limitations. Primarily, existing methods demonstrate insufficient robustness to intra-class variations within identical exposure categories, where images captured under disparate imaging conditions—including lighting environments, shooting contexts, and meteorological factors—may exhibit substantially different degradation characteristics. Even under equivalent exposure levels, degradation features induced by varying imaging conditions can differ significantly, rendering existing approaches inadequately adaptive to complex and dynamic real-world scenarios. Our empirical analysis through t-SNE\citep{van2008visualizing} clustering on the MSEC dataset (as shown in Fig. \ref{fig1} above) reveals substantial outliers that cluster closer to features from alternative exposure categories, resulting in degradation type misclassification and erroneous reconstruction strategies (illustrated in Fig. \ref{fig1} below). Furthermore, the majority of current methods employ end-to-end processing paradigms that optimize illumination recovery and detail reconstruction as unified objectives, leading to mutual interference between these inherently distinct processes and compromising overall correction performance.

To tackle these challenges, WEC-DG introduces an innovative multi-exposure correction framework guided by degradation descriptors. The degradation descriptors generated by the Scene Description Generation Module (SDGM) are utilized as conditional inputs in our approach, enabling explicit modeling and targeted processing of various degradation patterns. These descriptors are fed into the Exposure Consistency Alignment Module (ECAM), which is designed to ensure consistent alignment of exposures. By effectively capturing the degradation characteristics specific to different imaging conditions, these descriptors provide coarse-grained modulation signals that facilitate the transition from exposure-degraded images to normal exposure space. This design not only enhances the model's understanding of complex degradation patterns and its adaptability across diverse imaging scenarios but also contributes to more precise degradation recovery capabilities.

Additionally, through a thorough analysis of the frequency domain in multi-exposure images, we demonstrate that wavelet transforms can effectively decouple illumination from detail information components (see Fig. \ref{fig2}). Building on this insight, we developed a cascaded two-stage Exposure Restoration and Detail Reconstruction Module (EDRM). The initial stage focuses on processing low-frequency components to achieve precise illumination recovery, while the subsequent stage reconstructs details by utilizing high-frequency components based on the illumination-corrected intermediate results. This sequential processing strategy leverages the inherent properties of wavelet domain decomposition, ensuring accurate illumination correction while preserving high-quality detail recovery. As a result, it alleviates the common issues of illumination-detail interference typically associated with spatial domain processing. The primary contributions of this work are summarized as follows:
\begin{enumerate}
	\item A novel degradation descriptor-guided mechanism for multi-exposure enhancement. This mechanism effectively addresses the miscorrection of exposure anomalies caused by traditional methods' inability to recognize "blurred" exposure degradation, thus significantly improving the model’s generalizability and adaptability to complex imaging environments.
	\item A two-stage EDRM module leveraging wavelet transforms for illumination and detail processing. This module processes exposure and detail issues separately, ensuring accurate illumination correction while facilitating high-quality detail recovery. This sequential processing strategy effectively mitigates the interference issues prevalent in end-to-end frameworks.  
	\item Comprehensive experimental validation across multiple datasets. We conducted experiments on two exposure correction datasets and five zero-reference exposure anomaly datasets, demonstrating that our method outperforms state-of-the-art techniques in both objective metrics and visual quality, confirming its effectiveness and applicability.
\end{enumerate}

\section{Related Work}
\begin{figure}[t]
	\centering
	\includegraphics[width=1\columnwidth]{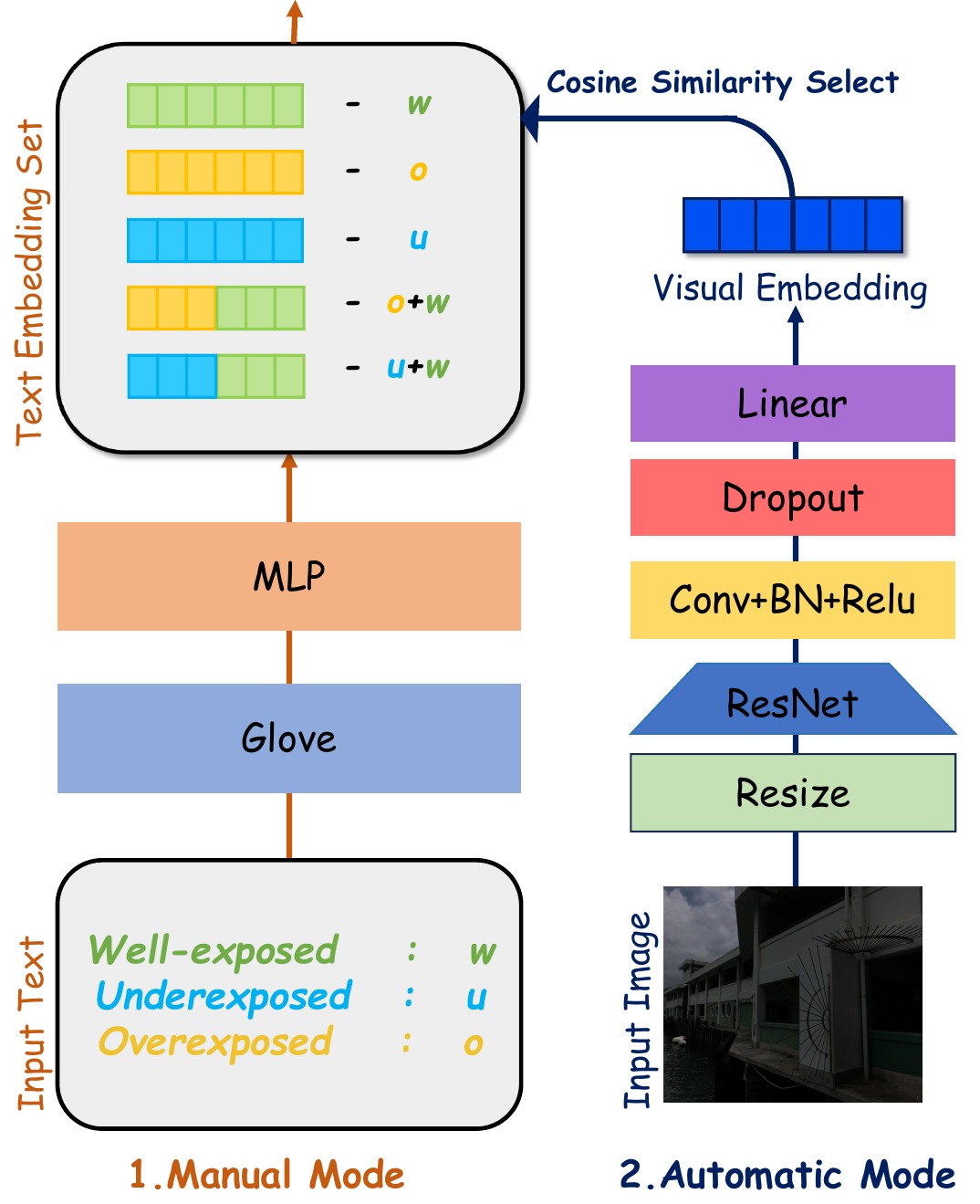} 
	\vspace{-1.5em}
	\caption{The structure of the Scene Description Generation Module (SDGM). This module enhances the controllability and adaptability of the image restoration model by operating in manual and automatic modes. }
	\vspace{-1.5em}
	\label{fig7}
\end{figure}
\subsection{Exposure Correction}
Image exposure correction has progressed from traditional algorithms to deep learning techniques. Early methods like histogram equalization \citep{gonzalez2002digital} and gamma correction \citep{poynton1996technical} are efficient but struggle with complex lighting, leading to over-enhancement. In contrast, deep learning methods such as RetinexNet \citep{wei2018deep} and Deep-UPE \citep{wang2019underexposed} utilize convolutional neural networks but face challenges of overfitting and limited generalization.

\begin{figure*}[t]  
	\centering  
	\vspace{-1em}
	\includegraphics[width=\textwidth]{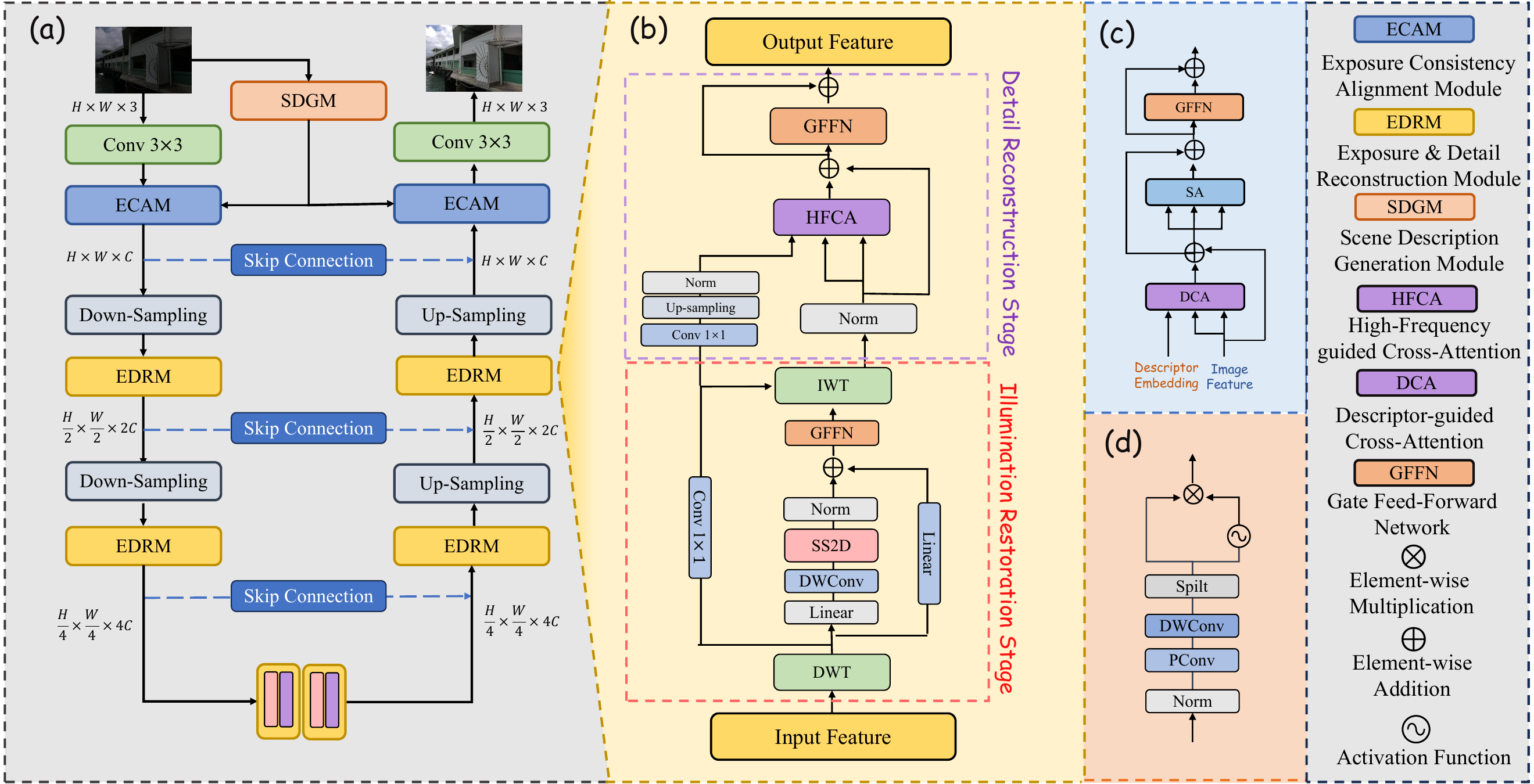}  
	\vspace{-1em}
	\caption{Illustration of the proposed model architecture: (a) Overall architecture of WEC-DG. (b) The Exposure Restoration and Detail Reconstruction Module (EDRM). (c) The Exposure Consistency Alignment Module (ECAM). (d) The Gate Feed-Forward Network.}  
	\vspace{-1em}
	\label{fig3}  
\end{figure*}

The focus has shifted to unified multi-exposure correction frameworks \citep{li2025osmamba, liu2024region}. MSEC \citep{afifi2021learning} uses a coarse-to-fine approach, while FECNet \citep{huang2022deep} addresses mixed exposure with local color distribution. Innovations like ERL \citep{huang2023learning} improve feature alignment. Recent methods, such as LACT \citep{baek2023luminance} and CoTF \citep{li2024real}, integrate global illumination modeling with local detail preservation. Despite advancements, challenges in addressing intra-class variations and interference between illumination and detail enhancement persist, necessitating better decoupling mechanisms.

\subsection{Wavelet Transform-Based Methods}
Wavelet transform is a powerful tool for frequency domain analysis in computer vision, particularly for illumination decoupling. Its integration with deep learning techniques has shown promise in image processing. For instance, \cite{liu2018multi} proposed a multi-level wavelet CNN that enhances receptive fields and balances efficiency with restoration. \cite{liu2020wavelet} developed a dual-branch network with spatial attention for effective moiré pattern removal. In super-resolution, \cite{xin2020wavelet} used wavelet decomposition to reconstruct high-resolution images, while \cite{li2024ewt} introduced an efficient wavelet Transformer for image denoising that reduces computational resources. Additionally, \cite{zou2024wave} applied wavelet transform to improve underexposed high-resolution images. However, existing methods often focus on task-specific architectures and do not thoroughly explore wavelet-based illumination recovery, especially in multi-exposure correction, highlighting the need for further investigation.

\section{Preliminaries}
\subsection{Discrete Wavelet Transform}  
In our framework, we utilize the two-dimensional discrete wavelet transform (DWT) \cite{daubechies1992ten} to decompose the input image in the frequency domain, effectively decoupling structural and detail information. For an input image $I \in \mathbb{R}^{H \times W \times C}$, DWT yields four subbands:  
\begin{equation}  
	\{c_A,c_H,c_V,c_D\}=\mathrm{DWT}(I)  
\end{equation}  
where $c_{A},c_{H},c_{V},c_{D} \in \mathbb{R}^{\frac{H}{2} \times \frac{W}{2} \times C}$ represent the low-frequency approximation and high-frequency detail components in horizontal, vertical, and diagonal directions, respectively. The low-frequency subband $c_{A}$ captures overall structure and illumination, while $c_{H},c_{V},c_{D}$ retain essential edges, textures, and fine details. This separation lays the groundwork for subsequent processing.  

The DWT is perfectly invertible; the subbands can be reconstructed into the original image via the inverse wavelet transform (IWT):  
\begin{equation}  
	I=\mathrm{I}\mathrm{W}\mathrm{T}(c_A,c_H,c_V,c_D)  
\end{equation}  
This ensures no information loss during frequency domain manipulations, preserving the fidelity of corrected images. 
\section{Overall Architecture}
\begin{table*}[htbp]  
	\centering   
	\resizebox{\textwidth}{!}{
		\begin{tabular}{>{\centering\arraybackslash}p{2cm}|>{\centering\arraybackslash}p{3cm}|cc|cc|cc|cc|cc|cc|c|c}  
			\hline  
			\multirow{3}{*}{\centering Methods} & \multirow{3}{*}{\centering Source} & \multicolumn{6}{c|}{MSEC}                     & \multicolumn{6}{c|}{SICE}                     & \multirow{3}{*}{Rank} & \multirow{3}{*}{RoR} \\
			\cline{3-14}          &       & \multicolumn{2}{c|}{under} & \multicolumn{2}{c|}{over} & \multicolumn{2}{c|}{average} & \multicolumn{2}{c|}{under} & \multicolumn{2}{c|}{over} & \multicolumn{2}{c|}{average} &       &  \\
			\cline{3-14}          &       & PSNR & SSIM  & PSNR & SSIM  & PSNR & SSIM  & PSNR & SSIM  & PSNR & SSIM  & PSNR & SSIM  &       &  \\
			\hline  
			RetinexNet  & BMVC 2018 & 12.13 & 0.6209 & 10.47 & 0.5953 & 11.30  & 0.6081 & 12.94 & 0.5171 & 12.87 & 0.5252 & 12.91 & 0.5212 & 14    & 13 \\
			SID   & CVPR 2018 & 19.37 & 0.8103 & 18.83 & 0.8055 & 19.10  & 0.8079 & 19.51 & 0.6635 & 16.79 & 0.6444 & 18.15 & 0.654 & 11.67 & 12 \\
			DRBN  & CVPR 2020 & 19.74 & 0.829 & 19.37 & 0.8321 & 19.56 & 0.8306 & 17.96 & 0.6767 & 17.33 & 0.6828 & 17.65 & 0.6798 & 10.58 & 10 \\
			Zero-DCE & CVPR 2020 & 14.55 & 0.5887 & 10.4  & 0.5142 & 12.48 & 0.5515 & 16.92 & 0.633 & 7.11  & 0.4292 & 12.02 & 0.5311 & 14.17 & 14 \\
			RUAS & CVPR 2021 & 13.43 & 0.6807 & 6.39  & 0.4655 & 9.91  & 0.5731 & 16.63 & 0.5589 & 4.54  & 0.3196 & 10.59 & 0.4393 & 15.08 & 16 \\
			SCI   & CVPR 2022 & 9.97  & 0.6681 & 5.83  & 0.519 & 7.90   & 0.5936 & 17.89 & 0.6401 & 4.45  & 0.3629 & 11.17 & 0.5015 & 14.75 & 15 \\
			MSEC  & CVPR 2021 & 20.52 & 0.8123 & 19.79 & 0.8156 & 20.16 & 0.814 & 19.62 & 0.6512 & 17.59 & 0.656 & 18.61 & 0.6536 & 10.67 & 11 \\
			ECLNet & ACM MM 2022 & 22.37 & 0.8566 & 22.7  & 0.8631 & 22.54 & 0.8599 & 22.05 & 0.6893 & 19.25 & 0.6872 & 20.65 & 0.6883 & 8.08 & 8 \\
			FECNet & ECCV 2022 & 22.96 & 0.8598 & 23.22 & 0.8748 & 23.09 & 0.8673 & 22.01 & 0.6737 & 19.91 & 0.6961 & 20.96 & 0.6849 & 6.92 & 7 \\
			MMHT  & ACM MM 2023 & 22.97 & 0.856 & 23.10  & 0.8709 & 23.04 & 0.8635 & 22.55 & 0.7090 & 21.06 & 0.7237 & 21.81 & \textcolor{darkgreen}{0.7164} & 5.33 & 5 \\
			CSEC  & CVPR 2024 & 22.18 & 0.8502 & 22.69 & 0.8662 & 22.44 & 0.8582 & 20.79 & 0.7031 & 20.02 & 0.7093 & 20.41 & 0.7062 & 8.08 & 8\\
			CoTF  & CVPR 2024 & \textcolor{blue}{23.36} & \textcolor{blue}{0.8630} & \textcolor{blue}{23.49} & \textcolor{blue}{0.8793} & \textcolor{blue}{23.43} & \textcolor{blue}{0.8712} & 22.90  & 0.7029 & 20.13 & \textcolor{darkgreen}{0.7274} & 21.52 & 0.7152 & 3.58 & 4 \\
			Wave-Mamba & ACM MM 2024 & 23.17 & 0.8606 & 22.73 & 0.8671 & 22.95 & 0.8639 & 23.12 & 0.7040 & 20.22 & 0.7102  & 21.67 & 0.707 & 5.42& 6 \\
			Exposure-slot  & CVPR 2025 & 23.09 & 0.8601 & 23.24 & \textcolor{darkgreen}{0.8762} & 23.17 & 0.8682 & \textcolor{blue}{23.85} & \textcolor{blue}{0.7092} & \textcolor{blue}{21.77} & \textcolor{blue}{0.7375} & \textcolor{blue}{22.81} & \textcolor{blue}{0.7234} & \textcolor{darkgreen}{3.25}  & \textcolor{darkgreen}{3} \\
			\hline  
			WEC-DG & \multirow{2}{*}{\centering Our Proposed} & \textcolor{darkgreen}{23.31} & \textcolor{blue}{0.8630} & \textcolor{darkgreen}{23.28} & 0.8756 & \textcolor{darkgreen}{23.30}  & \textcolor{darkgreen}{0.8693} & \textcolor{red}{24.25} & \textcolor{red}{0.7180} & \textcolor{darkgreen}{21.31} & 0.6970 & \textcolor{darkgreen}{22.78} & 0.7075 & \textcolor{blue}{3.17} & \textcolor{blue}{2} \\
			WEC-DG$^{\dagger}$  &        & \textcolor{red}{23.55} & \textcolor{red}{0.8656} & \textcolor{red}{23.62} & \textcolor{red}{0.8809} & \textcolor{red}{23.59} & \textcolor{red}{0.8733} & \textcolor{red}{24.25} & \textcolor{red}{0.7180} & \textcolor{red}{22.36} & \textcolor{red}{0.7391} & \textcolor{red}{23.31} & \textcolor{red}{0.7286} & \textcolor{red}{1}     & \textcolor{red}{1}  \\
			\hline  
		\end{tabular}%  
	}  
	\caption{Quantitative results on MSEC,  SICE are presented in terms of PSNR↑ and SSIM↑. The best scores are displayed in red, the second-best scores in blue, and the third-best scores in green. WEC-DG$^{\dagger}$ refers to the manual mode of the model, whereas WEC-DG denotes the automatic mode of the model.} 
	\vspace{-1.5em} 
	\label{tab1}%  
\end{table*}%  
The proposed method, WEC-DG, is illustrated in Fig. \ref{fig3}a and is structured as an encoder-decoder network, validated as an effective architectural paradigm for image restoration\citep{guo2023scanet, chen2023learning}. The architecture comprises three main components: the Scene Description Generation Module (SDGM), the Exposure Consistency Alignment Module (ECAM), and the Exposure Restoration and Detail Reconstruction Module (EDRM). WEC-DG employs a sequential, multi-stage workflow to refine the input image. It begins with the SDGM, which analyzes the degraded image to generate a degradation descriptor that captures the exposure artifacts. This descriptor then guides the ECAM in adjusting the image's exposure level, establishing a suitable baseline for further fine-tuning. Next, the image data undergoes multiple iterations through the EDRM, utilizing wavelet decomposition to optimize exposure and details. These modules iteratively correct residual errors and enhance texture. Finally, the feature maps are processed again by the ECAM to ensure global consistency, resulting in a visually coherent, high-quality restored image.
\subsection{Scene Description Generation Module (SDGM)}
The scene description generation module, inspired by \cite{guo2024onerestore}, adopts its architectural framework to enhance the controllability and adaptability of the image restoration model. It operates in dual modes: manual mode allows users to input scene description text, generating corresponding embeddings for precise exposure correction, while automatic mode extracts visual features from images and matches text embeddings using cosine similarity to enable large-scale processing. The text embedder relies on predefined degradation categories—underexposed, well-exposed, and overexposed—with initial embeddings derived from the GloVe methodology \cite{pennington2014glove}, which are refined using a multi-layer perceptron to accurately represent exposure characteristics. The visual embedder employs a ResNet-50 \cite{he2016deep} backbone to extract image features, processed through multiple steps to ensure alignment between visual and textual descriptions. Additional details about training details are available in the supplementary materials.
\subsection{Exposure Consistency Alignment Module (ECAM)}  
The Exposure Consistency Alignment Module (ECAM) is crucial in our degradation-description-guided adaptive multi-exposure wavelet correction method. As shown in Figure (c), ECAM effectively utilizes degradation descriptors from the Scene Degradation Generation Module and integrates three key components: the Degradation Context Aware module (DCA), Self-Attention mechanism (SA), and Gate Feed-Forward Network (GFFN) to enhance restoration.  

Our DCA employs degradation descriptor encoding as queries, represented mathematically as follows:  
\begin{equation}  
	DCA(Q_t, K, V) = \text{Softmax}\left(\frac{Q_t \cdot K^T}{\lambda}\right)V  
\end{equation}   
Here, \(\lambda\) is a temperature factor, \(Q_t\) represents the query matrix generated from scene descriptors, and \(K\) and \(V\) correspond to the key and value matrices derived from image features. The matrix \(Q_t\) is extracted from the input scene description embedding \(e_t\) through a linear transformation. To maintain token consistency during multiplication between \(Q_t\) and \(K\), we adjust the original image size before reshaping it to create \(K\). By integrating the SDCA module, each transformer block effectively combines scene descriptors and image features for targeted restoration.  

For the SA and GFFN modules, we adopt structures similar to those in \cite{zamir2022restormer}. The front segment of ECAM, positioned in the U-Net structure \cite{ronneberger2015u}, performs coarse-grained modulation of the overall exposure level, establishing a baseline for subsequent fine-tuning. The rear segment focuses on fine-grained exposure alignment, ensuring meticulous adjustments. By synchronizing these functions, ECAM enhances visual quality while maintaining global consistency across diverse imaging scenarios.  
\begin{figure*}[t]
	\centering
	\includegraphics[width=2\columnwidth]{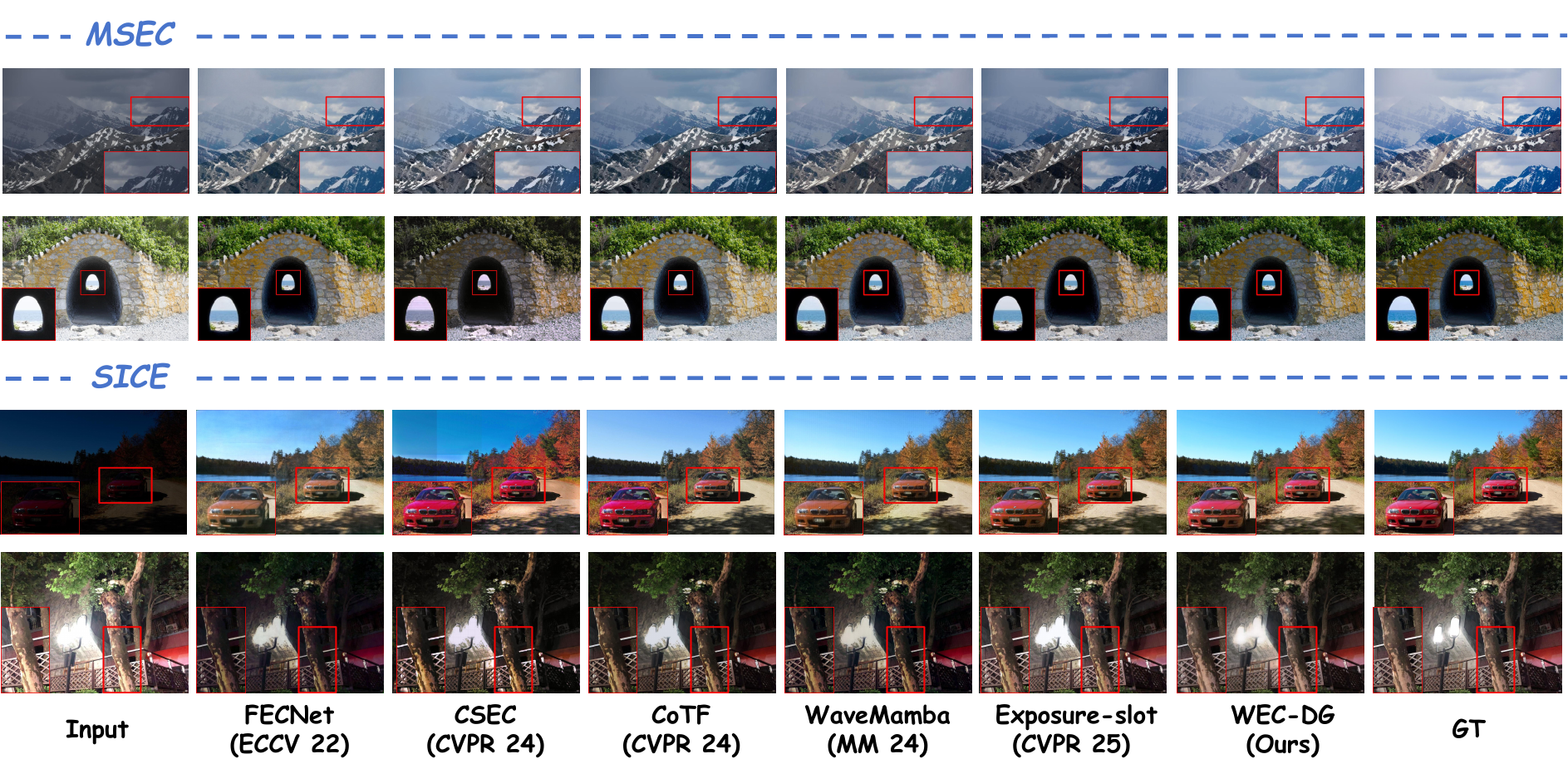} % Reduce the figure size so that it is slightly narrower than the column. Don't use precise values for figure width.This setup will avoid overfull boxes.
	\vspace{-0.5em}
	\caption{Visual comparison of our method against previous state-of-the-art approaches on MSEC and SICE datasets. In each dataset, the top row represents underexposed level, while the bottom row represents overexposed level.}
	\vspace{-1.2em}
	\label{fig4}
\end{figure*}
\subsection{Exposure Restoration and Detail Reconstruction Module (EDRM)}  
The Exposure Restoration and Detail Reconstruction Module (EDRM) addresses multi-exposure images in two stages: illumination restoration and detail reconstruction. In the Illumination Restoration Stage (see Fig. \ref{fig3}b), inspired by Mamba's achievements in modeling linear complex long-range dependencies, we employ discrete wavelet transform (DWT) and the VSSM to optimize low-frequency components. The Detail Reconstruction Stage employs high-frequency features with a high-frequency component-guided attention mechanism (HFCA) to reconstruct texture while minimizing artifacts. This interplay ensures the final output is rich and realistic.    
\subsubsection{Illumination Restoration Stage}  
In the Illumination Restoration Stage, our main goal is to enhance the image's low-frequency components. We achieve this by decomposing the image into low-frequency and high-frequency parts using discrete wavelet transforms. Guided by \cite{zou2024wave}'s findings on the efficiency of 2D Selective Scan Module 2D-SSM \cite{liu2024vmamba} for low-frequency processing in low-light image enhancement, for extracting and optimizing low-frequency global information, ultimately employing GFFN for feature integration and enhancement. This optimization process is mathematically represented as: 
In this stage, the input $X\in\mathbb{R}^{H\times W\times C}$ undergoes a Discrete Wavelet Transform (DWT) to generate low-frequency and high-frequency components, denoted as $X^L\in\mathbb{R}^{\frac{H}{2}\times \frac{W}{2}\times C}$ and $X^H\in\mathbb{R}^{\frac{H}{2}\times \frac{W}{2}\times 3C}$:  
\begin{equation}  
	X^L, X^H = \mathrm{DWT}(X)
\end{equation}  
Next, the low-frequency component $X^L$ begins with $\lambda C$ achieved using a linear layer, followed by activation with the SiLU function. Concurrently, a depth-wise convolution and SiLU activation function are applied in conjunction with a 2D Selective Scan Module (SS2D) and LayerNorm:  
\begin{equation}  
	X_1^L = LN\left(\mathrm{SS2D}\left(SiLU\left(DW\mathrm{Conv}\left(\mathrm{Linear}(X)\right)\right)\right)\right)  
\end{equation}  
Subsequently, \(X_1^L\) is combined with the original \(X^L\) using a linear transformation to produce \(X_2^L\):  
\begin{equation}  
	X_2^L = X_1^L + \text{Linear}(X^L)  
\end{equation}  

\begin{table}[t]  
	\centering    
	\begin{adjustbox}{max width=\columnwidth}  
		\begin{tabular}{ccccccc}  
			\hline  
			Methods & DICM  & LIME  & MEF   & NPE   & VV    & AVG \\
			\hline  
			FECNet & 2.52  & 1.68  & 1.66  & \textcolor{red}{2.04}  & 4.93  & 2.56 \\
			CSEC   & \textcolor{blue}{2.50}  & \textcolor{red}{1.56}  & 1.46  & 2.48  & 4.74  & \textcolor{blue}{2.54} \\
			CoTF   & 2.67  & \textcolor{blue}{1.58}  & 1.52  & 2.39  & 4.90  & 2.61 \\
			Wave-Mamba & 3.06  & 2.28  & \textcolor{red}{1.27}  & 2.32  & 5.14  & 2.81 \\
			Exposure-slot   & 2.73  & 2.10  & 1.61  & \textcolor{blue}{2.22}  & \textcolor{red}{4.05}  & \textcolor{blue}{2.54} \\
			\hline  
			WEC-DG   & \textcolor{red}{2.22}  & 1.80  & \textcolor{blue}{1.41}  & 2.42  & \textcolor{blue}{4.31}  & \textcolor{red}{2.43} \\
			\hline  
		\end{tabular}  
	\end{adjustbox}  
	\caption{RankIQA scores for LIME, VV, DICM, NPE, and MEF datasets. The best results are highlighted in red, while the second-best results are in blue. “AVG” represents the average RankIQA scores across these five datasets.}  
	\vspace{-1.5em} 
	\label{tab2}  
\end{table}
Then, a Gated Feature Fusion Network (GFFN) processes the feature representation \(X_1^L\), incorporating an added linear transformation to yield low-frequency features \(X_\text{en}^L\):  
\begin{equation}  
	X_\text{en}^L = \mathrm{GFFN}(X_1^L + \text{Linear}(X^L)) 
\end{equation}  
The GFFN leverages gating mechanisms to dynamically control the flow of information, enhancing the model's ability to capture complex patterns and relationships within the data. 
Finally, the enhanced image \(X_{\text{en}}\) is obtained through an Inverse Wavelet Transform (IWT):  
\begin{equation}  
	X_{\text{en}} = \mathrm{IWT}(X_\text{en}^L, \mathrm{Conv}(X^H))  
\end{equation}

\subsubsection{Detail Reconstruction Stage}  

In the Detail Reconstruction Stage, the high-frequency components extracted from the product of the wavelet transform in the Illumination Restoration Stage are utilized as prior guidance, enabling the model to focus more on the detailed texture information within the image through the high-frequency component-guided attention mechanism (HFCA).

First, the high-frequency components $X^H\in\mathbb{R}^{\frac{H}{2}\times \frac{W}{2}\times 3C}$ are adaptively adjusted using a \(1 \times 1\) convolution operating in the spatial domain. This is followed by applying a channel-wise \(1 \times 1\) convolution to achieve dimension alignment. Finally, up-sampling is performed to obtain details that match the pixel features of the exposure-enhanced image:
\begin{equation}  
	X_{prior}^{H}=LN\left(Up\left(\mathrm{Conv}\left(\mathrm{Conv}(X^{H})\right)\right)\right)  
\end{equation}   
These extracted details $X_{prior}$ are then fed into the HFCA attention mechanism, which is consistent with the ECAM structure (as shown in Fig \ref{fig3}c) to obtain detail-relevant residual features. The output is subsequently passed through the Gate Feed-Forward Network (GFFN) to achieve the final features $X_{out} $for exposure restoration and detail reconstruction. For simplicity, we have omitted the residual connections in the mathematical expression, which can be expressed as: 
\begin{equation}  
	X_{out} = \mathrm{GFFN}(\text{HFCA}(X_{prior}, X_{\text{en}})).  
\end{equation}

\section{Experiments}
\subsection{Experiment Settings}
\subsubsection{Implementation Details.}
Our method implementation utilizes PyTorch 1.12.0 and is trained a single NVIDIA 3090 GPU, using the PyTorch framework \cite{paszke2019pytorch}. The network contains 2.2M parameters. For optimization, we employ the Adam optimizer \cite{kingma2014adam} with exponential decay rates set to \( \beta_1 = 0.9 \) and \( \beta_2 = 0.999 \). Specifically, all images used for training are cropped into \( 512 \times 512 \) image patches with a sampling stride of 200. Additionally, the images are randomly flipped at angles of 0, 90, 180, and 270 degrees.
\subsubsection{Loss Function.}  
We use $\mathrm{L}$1 loss $\mathcal{L}_{1}$, SSIM loss $\mathcal{L}_{ssim}$\cite{wang2004image}, contrastive loss $\mathcal{L}_{con}$\cite{guo2024onerestore}, and perceptual loss $\mathcal{L}_{per}$\cite{johnson2016perceptual} to train our model. The overall loss is represented as:  

\begin{equation}  
	\mathcal{L} = \lambda_{1} \mathcal{L}_{1} + \lambda_{2} \mathcal{L}_{ssim} + \lambda_{3} \mathcal{L}_{con} + \lambda_{4} \mathcal{L}_{per}  
\end{equation}  

where the weights for each loss term are set as $\lambda_{1}=0.7$, $\lambda_{2}=0.3$, $\lambda_{3}=0.1$, and $\lambda_{4}=0.3$.
\subsubsection{Datasets and Evaluation metrics.}
We assess the performance of our method using several representative datasets: the multi-exposure dataset MSEC \cite{afifi2021learning}, the SICE dataset \cite{cai2018learning}, and five zero-reference exposure abnormal datasets: LIME (10 images) \cite{lee2013contrast}, MEF (17 images) \cite{guo2016lime}, VV (24 images) \cite{ma2015perceptual}, NPE (85 images) \cite{wang2013naturalness}, and DICM (64 images) \cite{yang2020advancing}. The MSEC dataset contains images depicting two levels of under-exposure and three levels of over-exposure for each scene. In our evaluation, we use 17,675 images for training and 5,905 images for testing purposes. The SICE dataset features both under- and over-exposure levels, with the train-test split following the approach outlined in \cite{huang2022exposure}. Evaluations are conducted using Peak Signal-to-Noise Ratio (PSNR) and Structural Similarity Index (SSIM) \cite{wang2004image} metrics. Additionally, we adopt the methodology from \cite{li2023pixel}, using the average ranking of various metrics for each dataset, referred to as “Rank.” We also compute the rank of these “Rank” values, termed “RoR,” to ensure a fair comparison across multiple metrics. As for zero-reference datasets, we employ RankIQA \cite{liu2017rankiqa} as an unsupervised quality assessment method.
\subsection{Comparison with State-of-the-Art Methods}  

We compare the proposed WEC-DG method with seven low-light enhancement methods, namely RetinexNet \cite{wei2018deep}, SID \cite{chen2018learning}, DRBN \cite{yang2020fidelity}, Zero-DCE \cite{guo2020zero}, RUAS \cite{liu2021retinex}, SCI \cite{ma2022toward} and Wave-Mamba \cite{zou2024wave}, as well as seven state-of-the-art multi-exposure correction techniques: MSEC \cite{afifi2021learning}, ECLNet \cite{huang2022exposure}, FECNet \cite{huang2022deep}, MMHT \cite{li2023fearless}, CSEC \cite{li2024color}, CoTF \cite{li2024real} and Exposure-slot \cite{jung2025exposure}.
\subsubsection{Quantitative Comparisons.}
Tab. \ref{tab1} presents performance metrics, including PSNR and SSIM, across the MSEC and SICE datasets. The results indicate that WEC-DG outperforms these existing methods, achieving a PSNR of 23.59 and an SSIM of 0.8733 on MSEC, and a PSNR of 23.31 and an SSIM of 0.7286 on SICE. These findings highlight WEC-DG's robustness and efficiency in handling varying exposure levels, showcasing its potential for real-world image processing applications. In addition, we achieved the highest average RankIQA score on the zero-reference exposure anomaly dataset, reflecting the strong generalization capability of our WEC-DG method. Furthermore, as illustrated in Tab. \ref{tab2}, our WEC-DG method attained the highest average RankIQA score on the zero-reference exposure anomaly dataset, highlighting its robust generalization capability.
\subsubsection{Qualitative Comparisons.}
Fig. \ref{fig4} shows a qualitative comparison of our WEC-DG method and other approaches. In the MSEC dataset, other methods exhibit noticeable color shifts during image processing, as highlighted in the red box in the Fig. \ref{fig4}. On the SICE dataset, CSEC yields prominent spurious boundaries in underexposed images, while both FECNet and WaveMamba show a dull recovery of colors, as illustrated by the red sports car in the figure. In contrast, our method effectively restores authentic details and textures in overexposed images, particularly evident in the bark of the tree shown in the image. This demonstrates our approach's capability to maintain color fidelity and detail integrity across varying exposure conditions. The visualization of the zero-reference exposure anomaly dataset can be found in the supplementary materials.

\begin{table}[t]  
	\centering    
	\begin{adjustbox}{max width=\columnwidth}  
		\begin{tabular}{c|ccccc|cc}  
			\hline  
			\multirow{2}{*}{Settings} & \multirow{2}{*}{ECAM} & \multirow{2}{*}{IRS} & \multirow{2}{*}{DRS} & \multirow{2}{*}{$\mathcal{L}_{per}$} & \multirow{2}{*}{$\mathcal{L}_{con}$} & \multirow{2}{*}{PSNR} & \multirow{2}{*}{SSIM} \\  
			&       &       &       &       &       &       &  \\  
			\hline  
			a     &       & \textcolor{darkgreen}{\checkmark} & \textcolor{darkgreen}{\checkmark} & \textcolor{darkgreen}{\checkmark} & \textcolor{darkgreen}{\checkmark} & 21.23 & 0.7036 \\  
			b     & \textcolor{darkgreen}{\checkmark} &       & \textcolor{darkgreen}{\checkmark} & \textcolor{darkgreen}{\checkmark} & \textcolor{darkgreen}{\checkmark} & 22.49 & 0.7112 \\  
			c     & \textcolor{darkgreen}{\checkmark} & \textcolor{darkgreen}{\checkmark} &       & \textcolor{darkgreen}{\checkmark} & \textcolor{darkgreen}{\checkmark} & 22.77 & 0.7132 \\  
			d     & \textcolor{darkgreen}{\checkmark} & \textcolor{darkgreen}{\checkmark} & \textcolor{darkgreen}{\checkmark} &       & \textcolor{darkgreen}{\checkmark} & 21.11 & 0.6979 \\  
			e     & \textcolor{darkgreen}{\checkmark} & \textcolor{darkgreen}{\checkmark} & \textcolor{darkgreen}{\checkmark} & \textcolor{darkgreen}{\checkmark} &       & \textcolor{blue}{23.12} & \textcolor{blue}{0.7213} \\  
			\hline  
			Ours  & \textcolor{darkgreen}{\checkmark} & \textcolor{darkgreen}{\checkmark} & \textcolor{darkgreen}{\checkmark} & \textcolor{darkgreen}{\checkmark} & \textcolor{darkgreen}{\checkmark} & \textcolor{red}{23.31} & \textcolor{red}{0.7286} \\  
			\hline  
		\end{tabular}  
	\end{adjustbox}  
	\caption{Ablation studies on proposed modules and losses.}  
	\vspace{-1em}
	\label{tab3}  
\end{table}
\subsection{Ablation Study}
This section examines the impact of various modules on the proposed model's performance. As illustrated in Tab. \ref{tab3}, in Setting a, we replaced the degradation-guided attention mechanism DCA in the Exposure Consistency Alignment Module (ECAM) with the self-attention mechanism CA to evaluate the effectiveness of the degradation descriptor for accurate exposure alignment. The results reveal a significant decline in recovery capability without DCA, underscoring its essential role in managing image degradation. In Settings b and c, we substituted the Illumination Restoration Stage (IRS) and Detail Reconstruction Stage (DRS) of the Exposure Restoration and Detail Reconstruction Module (EDRM) with transformer blocks, allowing us to assess their performance in exposure restoration and validate our wavelet transform-based processing module. These modifications improved the model's handling of image restoration tasks and overall quality. Additionally, ablation studies on the loss functions \(L_{per}\) and \(L_{con}\) show that incorporating perceptual loss enhances image quality and detail reproduction, resulting in better recovery capability. Contrastive loss further improves performance and generalization by reducing feature coupling across categories. These findings highlight the importance of combining diverse loss functions and modular components for effective image restoration.

\begin{figure}[t]
	\centering
	\includegraphics[width=1\columnwidth]{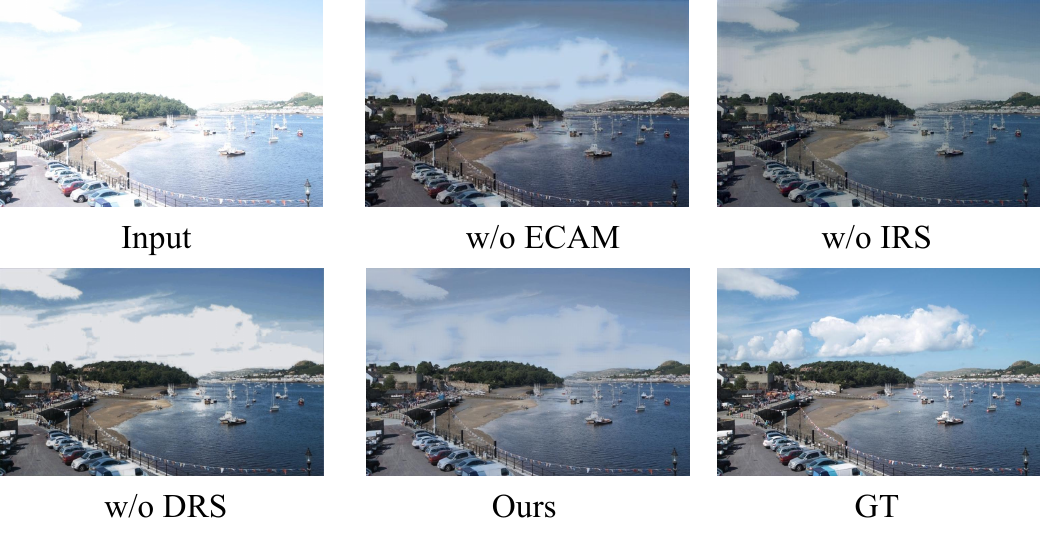} % Reduce the figure size so that it is slightly narrower than the column. Don't use precise values for figure width.This setup will avoid overfull boxes.
	\caption{The visualization of the ablation study.}
	\vspace{-1em}
	\label{fig5}
\end{figure}

The visualization of the ablation study results, as shown in Fig. \ref{fig5}, illustrates that when the ECAM does not function, the model is prone to excessive recovery, resulting in darkened images. Furthermore, when the Illumination Restoration Stage (IRS) is inactive, accurate exposure corrections cannot be achieved. Similarly, the absence of the Detail Reconstruction Stage (DRS) leads to severe distortion of image details, particularly in cloud formations.

\section{Conclusion}  
This study proposes WEC-DG, a wavelet-based exposure correction method that combines a degradation guidance mechanism and a two-stage recovery module to address multi-exposure image enhancement challenges. By utilizing carefully designed degradation descriptors, this method can effectively detect and correct "blurred" exposure degradation issues that traditional methods often fail to handle, thereby significantly improving the model's generalization capability across various complex imaging environments. The core Exposure Restoration and Detail Reconstruction Module employs discrete wavelet transforms, strategically decoupling lighting processing from detail processing while effectively mitigating common interference problems inherent in traditional end-to-end frameworks. Comprehensive experiments conducted on multiple benchmark datasets validate the effectiveness of this method, demonstrating superior performance that significantly surpasses existing state-of-the-art techniques in both quantitative metrics and perceptual visual quality.

\bibliography{aaai2026}

\section{Supplementary Explanation for Scene Description Generation Module}
\subsection{Setting of Degradation Descriptors}
In our study, we utilize two distinct datasets: SICE and MSEC, each with specific configurations. Initially, we define three fundamental descriptive terms: "underexposed," "well-exposed," and "overexposed," based on pre-trained GloVe word vectors from Twitter. For the SICE dataset, these terms correspond to the settings "under," "over," and "GT," respectively. Conversely, the MSEC dataset features two degrees of overexposure (P1.5 and P1) and two degrees of underexposure (N1.5 and N1). In this case, we assign "underexposed" to N1.5 and "overexposed" to P1.5, while the term "well-exposed" corresponds to "GT." For the levels N1 and P1, we calculate a weighted average of the descriptive word vectors for "underexposed" and "well-exposed," as well as for "overexposed" and "well-exposed," to accurately represent these degradation levels.

\subsection{Training Details}  

We utilized the Onerestore training approach to develop our Scene Description Generation Module (SDGM). In this setup, we specified a learning rate of \(1 \times 10^{-4}\), trained for 100 epochs, and employed a batch size of 128.  

To compute the similarity scores between the visual embeddings \(e_v\) and text embeddings \(e_t\), we adopted a Cosine Cross-Entropy method, as referenced in [62]. The similarity score \(S(e_v, e_t)\) is formulated as follows:  

\begin{equation}  
	\cos(e_v, e_t) = \delta \cdot \frac{e_v \cdot e_t^T}{\| e_v \| \| e_t \| }  
\end{equation}  

where \(\delta\) is a scaling factor. The overall similarity score is expressed as:  

\begin{equation}  
	S(e_v, e_t) = \frac{\sum_{i=1}^{N_t} e^{\cos(e_{v}, e_{t_i})}}{\sum_{j=1}^{N_t} e^{\cos(e_{v}, e_{t_j})}}  
\end{equation}  

Here, \(N_t\) denotes the total number of text embeddings. Using the computed similarity score \(S(e_v, e_t)\), we applied a cross-entropy loss function for training the model. This method effectively facilitates the alignment between visual and textual representations, contributing to accurate scene description generation.

\section{Visualization Results of Zero-Reference Exposure Anomaly Datasets}
Figures \ref{fig8} and \ref{fig9} present the visual results of our proposed method, WEC-DG, compared with six other state-of-the-art approaches across five zero-reference exposure anomaly datasets. All methods were trained on SICE to ensure a fair comparison. Below, we provide a qualitative analysis of the performance of each method on the different datasets.

\subsection{DICM:} 
\begin{figure*}
	\centering
	\includegraphics[width=2\columnwidth]{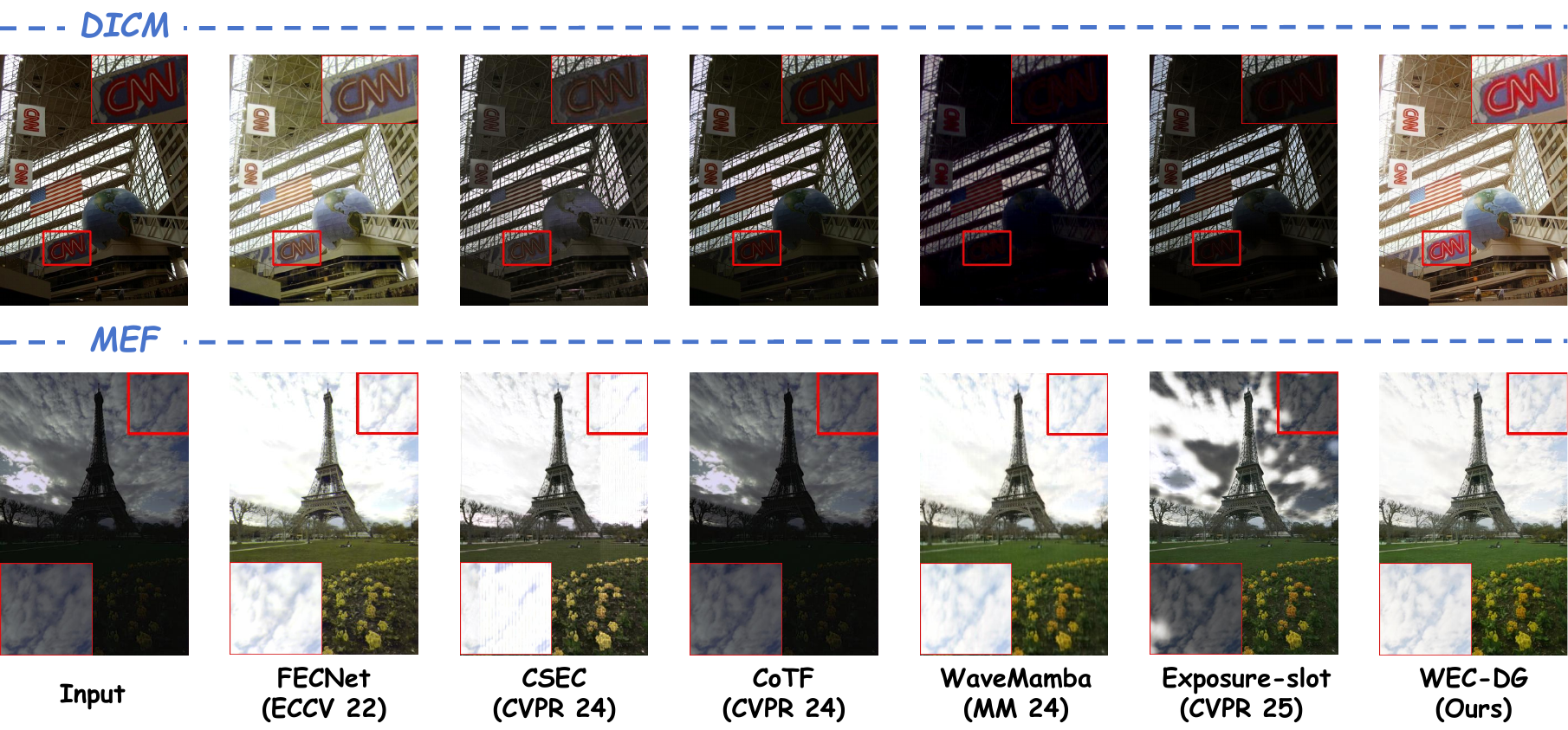} % Reduce the figure size so that it is slightly narrower than the column. Don't use precise values for figure width.This setup will avoid overfull boxes.
	\caption{Visual comparison of DICM and MEF datasets. Zoom in to Examine Details.}
	\label{fig8}
\end{figure*}

This is a typical example of erroneous recovery, as demonstrated in Figure \ref{fig8}. The methods CSEC, CoTF, WaveMamba, and Exposure-slot failed to accurately restore exposure towards the desired enhancement, resulting in even more severe exposure degradation. While FECNet managed to recover normal exposure, it led to noticeable color degradation, as highlighted by the red box in the figure. In contrast, our method not only restored exposure but also preserved more realistic color information.

\subsubsection{MEF:} In this instance, CoTF did not correctly enhance the exposure of the image, while Exposure-slot introduced significant artifacts during the recovery process. Among the methods that successfully restored the image, FECNet exhibited color biases, particularly evident in the color of the grass, while CSEC created noticeable brightness boundaries. Our method ensured accurate color representation and realistic texture details during restoration, as evidenced by the cloud details within the red box.

\subsubsection{LIME:} 
\begin{figure*}
	\centering
	\vspace{-10em}
	\includegraphics[width=2\columnwidth]{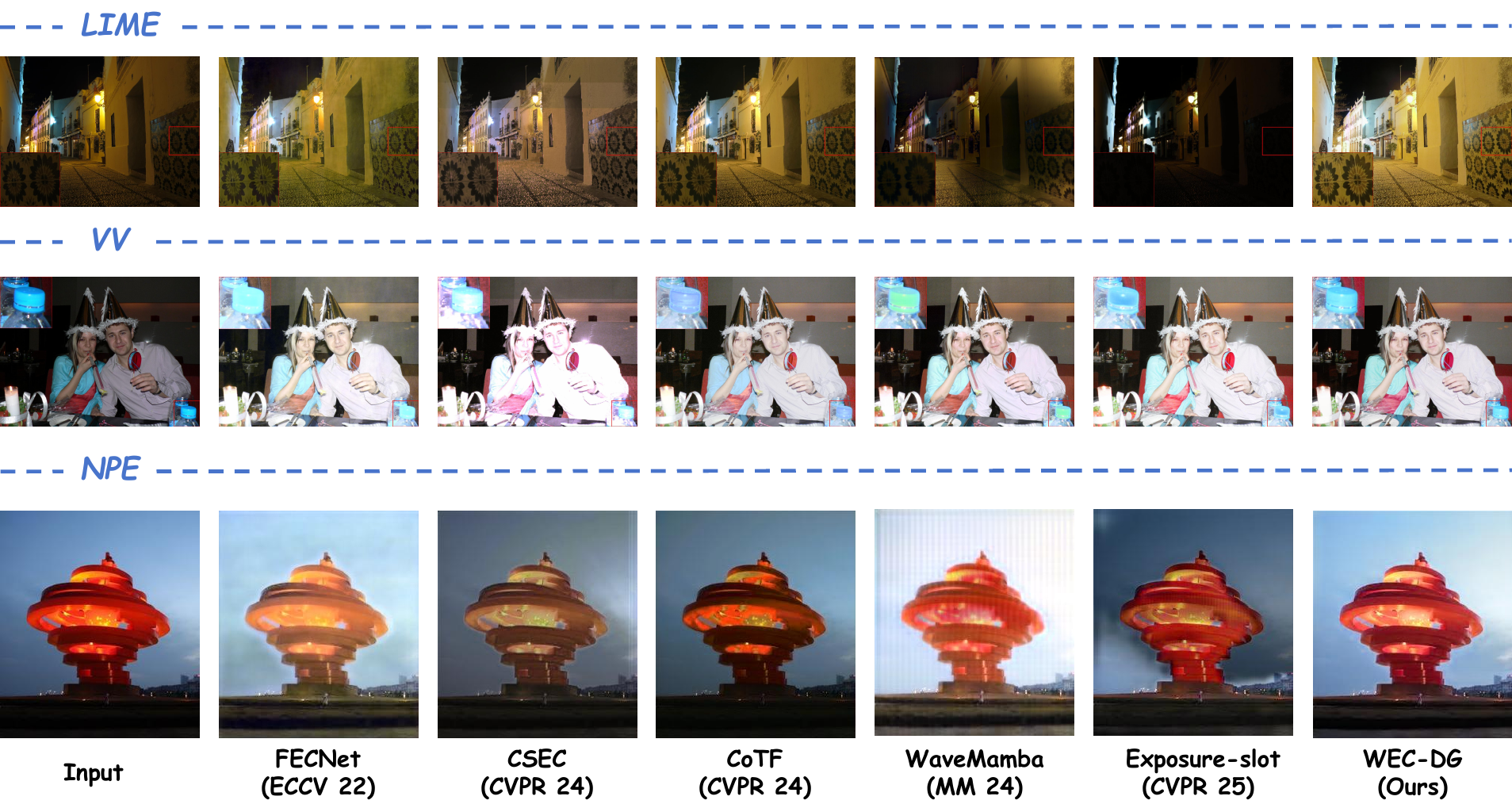} % Reduce the figure size so that it is slightly narrower than the column. Don't use precise values for figure width.This setup will avoid overfull boxes.
	\caption{Visual comparison of DICM and MEF datasets. Zoom in to Examine Details.}
	\label{fig9}
\end{figure*}

Both WaveMamba and Exposure-slot introduced correction errors during processing, while FECNet, CSEC, and CoTF did not achieve the expected levels of normal exposure restoration, as shown in the red box of Figure \ref{fig9}.

\subsubsection{VV:} In this example, all methods succeeded in enhancing the image to normal exposure levels. However, in managing image details—specifically in the bottle cap area highlighted by the red box in Figure 2—FECNet resulted in excessive enhancement, while CSEC and WaveMamba exhibited color biases. CoTF and Exposure-slot introduced varying degrees of artifacts.

\subsubsection{NPE:} FECNet and CSEC both suffered from different levels of color bias, whereas COTF's recovery was insufficient. WaveMamba exhibited over-recovery, and Exposure-slot resulted in visible artifacts.
\section{Exposure Consistency Alignment Module (ECAM) as a Plug-and-Play Module}
\begin{figure*}
	\centering
	\includegraphics[width=2\columnwidth]{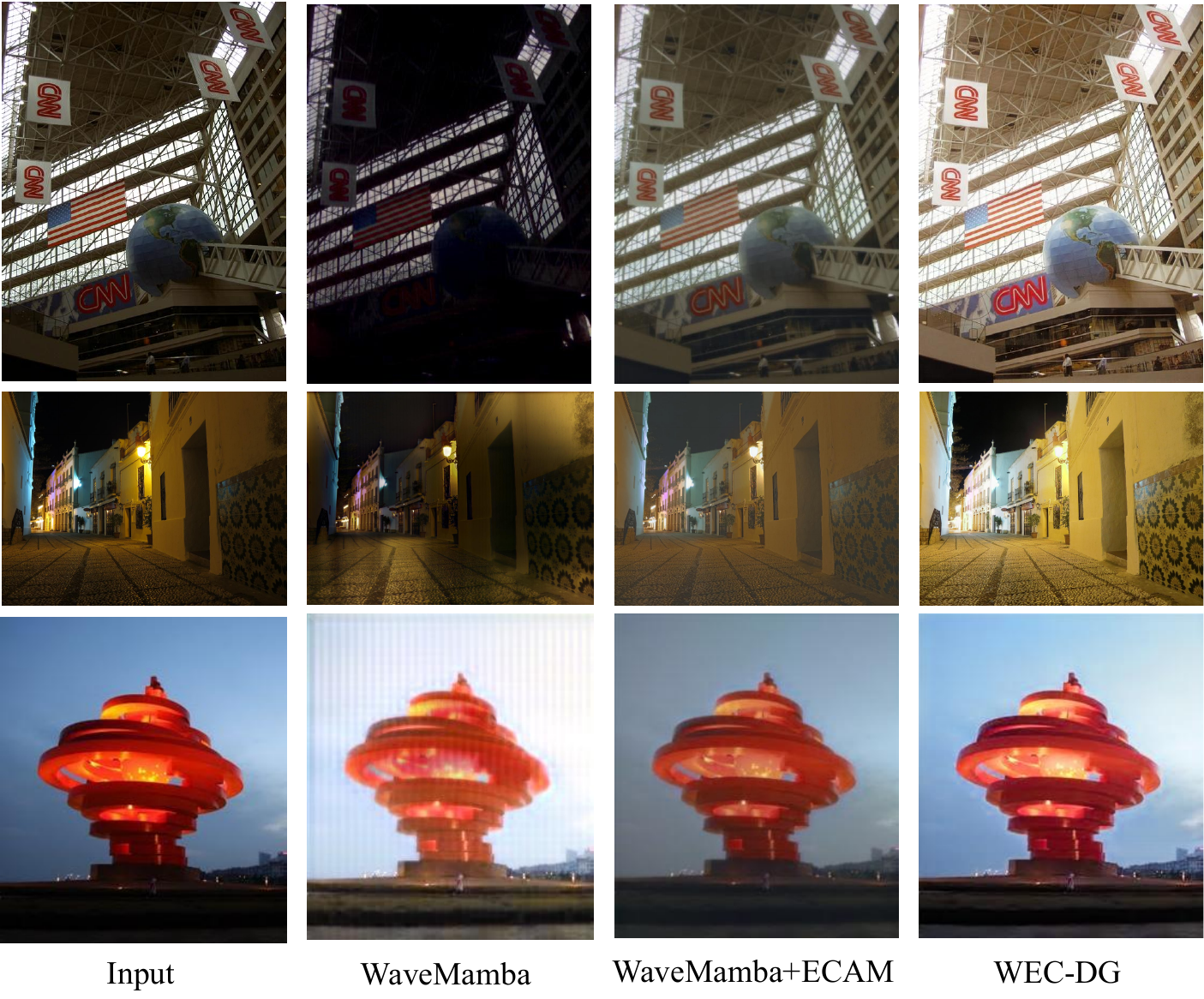} % Reduce the figure size so that it is slightly narrower than the column. Don't use precise values for figure width.This setup will avoid overfull boxes.
	\caption{ECAM provides WaveMamba with the ability to accurately process degradation.}
	\label{fig10}
\end{figure*}

As shown in Table \ref{tab4}, the results indicate that after the inclusion of ECAM, WaveMamba exhibits significant improvements in both Peak Signal-to-Noise Ratio (PSNR) and Structural Similarity Index (SSIM) metrics on the SICE dataset. These enhancements suggest that ECAM plays a crucial role in optimizing the model's performance by refining the quality of the images produced.
In addition, Figure \ref{fig10} illustrates the findings from re-evaluating WaveMamba’s error handling capabilities on the DICM, LIME, and NPE datasets. Through the application of SDGM, the model demonstrated its proficiency in accurately identifying image degradation. Furthermore, under the modulation of ECAM, WaveMamba successfully restored the degraded images to a level of clarity and quality that underscores the effectiveness of the ECAM integration. 
\begin{table}[t]  
	\centering  
	\begin{adjustbox}{max width=\columnwidth}  
		\begin{tabular}{l|rr|rr|rr}  
			\hline  
			\multicolumn{1}{c|}{\multirow{2}[4]{*}{Models}} & \multicolumn{2}{c|}{Underexposed} & \multicolumn{2}{c|}{Overexposed} & \multicolumn{2}{c}{Average} \\
			\cline{2-7}  
			& PSNR & SSIM & PSNR & SSIM & PSNR & SSIM \\
			\hline  
			WaveMamba & 23.12 & 0.704  & 20.22 & 0.710 & 21.67 & 0.707 \\
			WaveMamba+ECAM & 23.47 & 0.708  & 20.78 & 0.714 & 22.13 & 0.711 \\
			\hline  
		\end{tabular}  
	\end{adjustbox}  
	\caption{Performance metrics for WaveMamba under different exposure conditions.}  
	\vspace{-1.5em}  
	\label{tab4}  
\end{table}

\end{document}